\documentclass{article} 
\usepackage[preprint]{care_preprint}
\usepackage{microtype}

\usepackage{amsmath,amsfonts,bm}

\def\eqref#1{equation~\ref{#1}}

\def\1{\bm{1}}

\DeclareMathAlphabet{\mathsfit}{\encodingdefault}{\sfdefault}{m}{sl}
\SetMathAlphabet{\mathsfit}{bold}{\encodingdefault}{\sfdefault}{bx}{n}

\usepackage{graphicx}
\usepackage{subcaption}
\usepackage{wrapfig}
\usepackage{hyperref}
\usepackage{url}
\definecolor{darkblue}{rgb}{0,0,0.5}
\hypersetup{colorlinks=true,citecolor=darkblue,linkcolor=darkblue,urlcolor=darkblue}
\usepackage{algorithm}
\usepackage{algpseudocode}
\usepackage{amssymb}
\usepackage{booktabs}
\usepackage{multirow}
\usepackage[table]{xcolor}
\usepackage{makecell}
\usepackage{tabularx}

\definecolor{carebg}{HTML}{E9EEF8}
\definecolor{careblue}{HTML}{4F7DF3}
\definecolor{careheat}{RGB}{70,172,163}
\definecolor{appendixheader}{RGB}{205,202,238} 
\definecolor{appendixpurple}{RGB}{238,237,249} 
\newcommand{\carecell}[1]{\cellcolor{carebg}#1}
\definecolor{gainblue}{RGB}{60,110,170}
\newcommand{\gain}[1]{\textcolor{gainblue}{\scriptsize $(+#1)$}}

\title{Mitigating Cross-Modal Hallucinations by Correcting Modality Misrouting in Omnimodal LLMs}

\author{%
  \textbf{Yu Mi}\textsuperscript{1,2},
  \textbf{Licheng Pan}\textsuperscript{1},
  \textbf{Jiayi Zhu}\textsuperscript{1},
  \textbf{Qirui Liu}\textsuperscript{1},
  \textbf{Yitong Qiao}\textsuperscript{1}\\[3pt]
  \textbf{Changtong Xu}\textsuperscript{2},
  \textbf{Tao Wei}\textsuperscript{2},
  \textbf{Fei Sun}\textsuperscript{3},
  \textbf{Zhixuan Chu}\textsuperscript{1}\thanks{Corresponding author.},
  \textbf{Kui Ren}\textsuperscript{1}\\[5pt]
  {\small
    \textsuperscript{1}Zhejiang University\quad
    \textsuperscript{2}Ant Group\quad
    \textsuperscript{3}Institute of Computing Technology, CAS}%
}

\renewcommand{\headrulewidth}{1.5pt}
\fancypagestyle{plain}{%
  \fancyhead{}
  \fancyhead[L]{\small Preprint}
  \renewcommand{\headrulewidth}{1.5pt}
}
\fancypagestyle{firstpage}{%
  \fancyhead{}
  \renewcommand{\headrulewidth}{0pt}
}

\makeatletter
\renewcommand{\@maketitle}{%
  \noindent
  \includegraphics[height=0.36in]{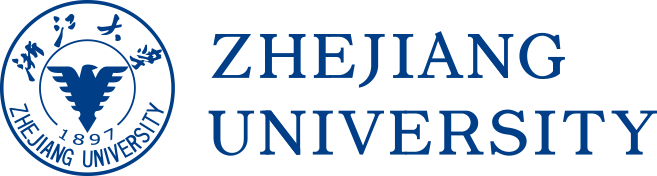}%
  \hspace{1.5em}%
  \includegraphics[height=0.38in]{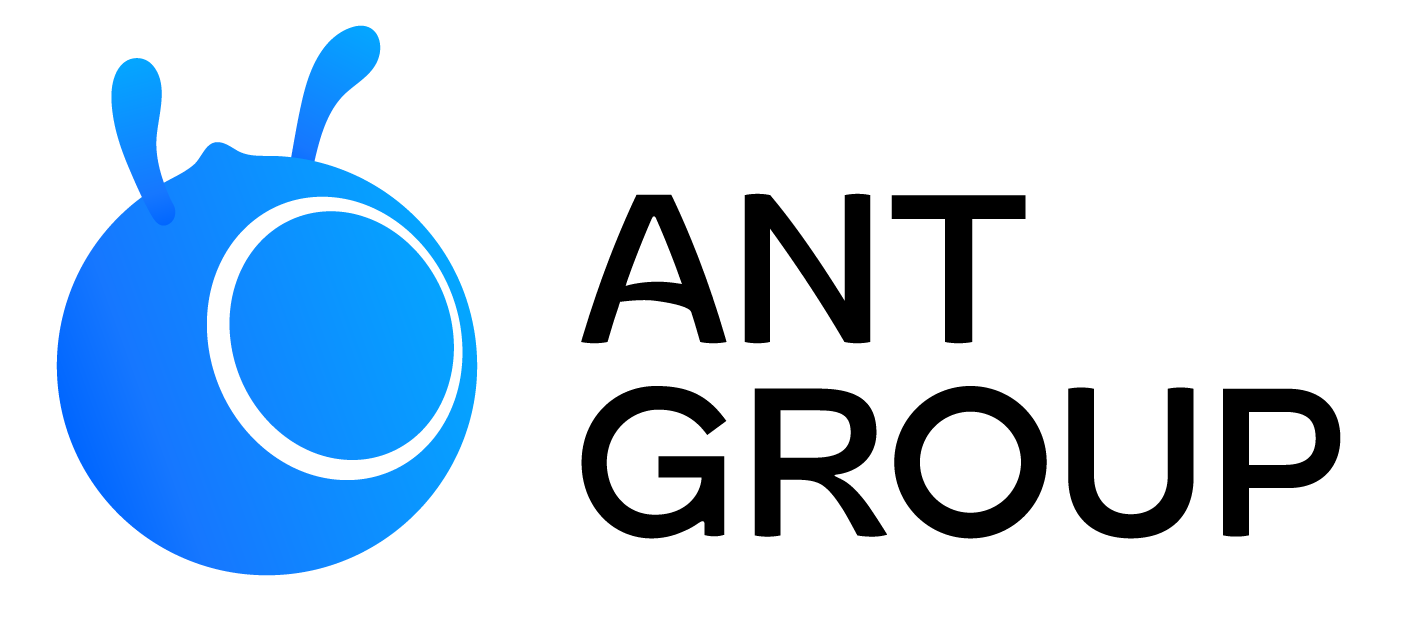}%
  \par
  \vspace{0.2em}
  \nointerlineskip
  \noindent\rule{\textwidth}{1.5pt}\par
  \vspace{0.6em}
  \begin{center}
    {\Large\bfseries \@title\par}
    \vspace{1.5em}
    {\normalsize \@author\par}
  \end{center}
  \vspace{0.8em}
}
\makeatother

\begin{document}
\pagestyle{fancy}
\maketitle
\thispagestyle{firstpage}

\begin{abstract}
Omnimodal large language models (OLLMs) are susceptible to cross-modal hallucinations under audio-visual interference. We find that, for a substantial fraction of incorrect joint predictions, the question-relevant unimodal branch can still recover the correct answer, suggesting that task-relevant evidence can remain recoverable from the relevant modality even when it is not reflected in the joint prediction.
We refer to this recoverable failure pattern as \emph{modality misrouting}. We propose \textbf{CaRe (Conflict-aware Residual Intervention)}, a training-free inference-time method for identifying and correcting modality misrouting. CaRe uses a text-only modality probe to identify a candidate modality-specific counterfactual branch, and flags potential misrouting when its prediction conflicts with the original joint prediction. For such cases, CaRe computes a residual direction from the discrepancy between the counterfactual and fused representations and uses it to steer the original multimodal state toward the relevant modality evidence, while retaining both modalities in the original audio-visual inference path. The intervention strength is adapted based on the confidence of the counterfactual branch. Experiments on CMM and AVHBench across two OLLMs show that CaRe consistently outperforms the evaluated inference-time hallucination-mitigation baselines.
\end{abstract}

\section{Introduction}

Omnimodal large language models (OLLMs) jointly process audio, visual, and textual signals, enabling cross-modal perception and reasoning \citep{Qwen2.5-Omni,minicpm,videollama,vita,openomni}. However, joint multimodal modeling does not necessarily ensure faithful grounding. OLLMs may still produce responses that are not faithfully grounded in the available audio-visual evidence, leading to audio-visual hallucinations that limit their reliability in video understanding, audio-visual question answering, and embodied applications \citep{cmm,sung2025avhbench}.

\begin{figure}[t]
    \centering
    \includegraphics[width=\textwidth]{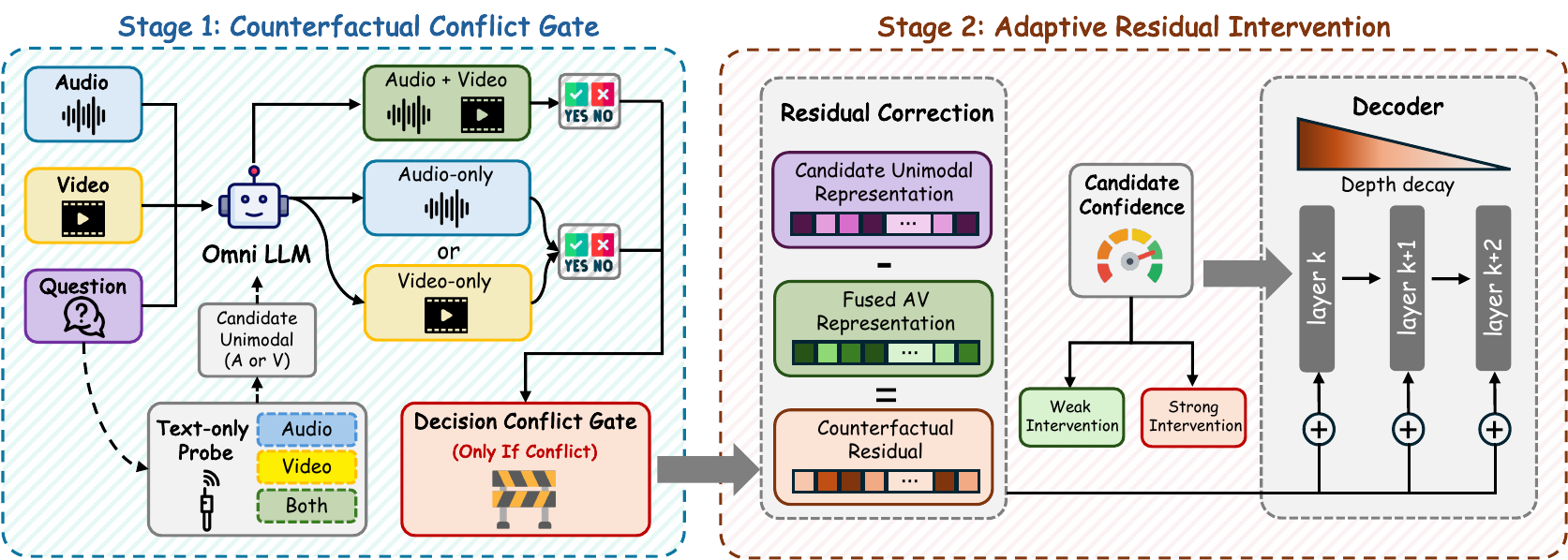}
    \caption{\textbf{Overview of CaRe.} Given audio-visual inputs and a question, CaRe uses a text-only modality probe to select a candidate modality, detects decision conflicts between the joint and counterfactual branches, and applies confidence-adaptive residual intervention at selected decoder layers with strengths decayed across depth.}
    \label{fig:care-pipeline}
\end{figure}

\begin{wrapfigure}{r}{0.6\linewidth}
    \centering
    \vspace{-1.3em}
    \includegraphics[width=\linewidth]{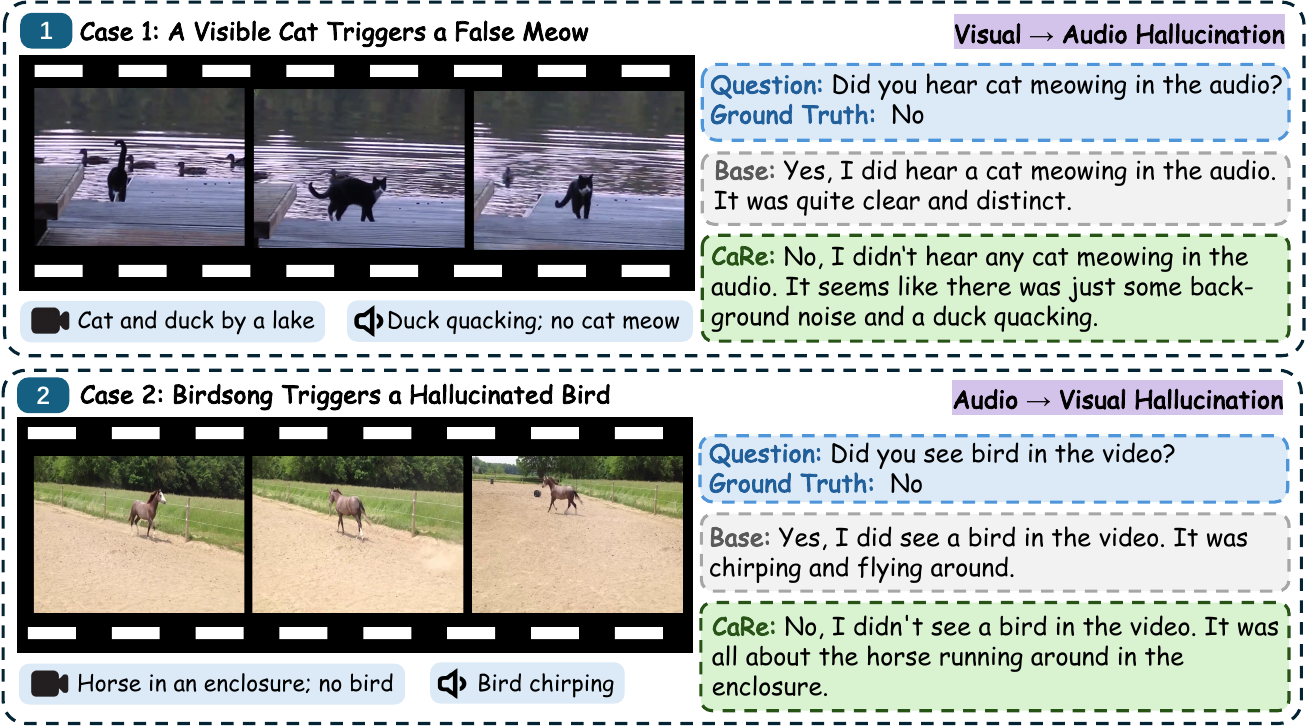}
    \vspace{-1.5em}
    \caption{\textbf{Bidirectional modality misrouting under audio-visual interference.} CaRe mitigates these errors by steering the fused representation toward the selected modality-specific branch.}
    \label{fig:cases}
    \vspace{-1.5em}
\end{wrapfigure}

We investigate a recoverable form of cross-modal hallucination in which one modality interferes with evidence from the modality actually required by the question. As illustrated in Figure~\ref{fig:cases}, a visible cat can cause an OLLM to falsely report hearing a meow even when the audio contains only duck quacking; conversely, birdsong can induce a hallucinated bird in the visual scene. Importantly, these errors do not necessarily imply that the question-relevant evidence has been lost: for a substantial fraction of incorrect joint predictions, the corresponding unimodal branch still recovers the correct answer. We refer to this discrepancy between the erroneous joint prediction and the recoverable question-relevant modality-specific prediction as \emph{modality misrouting}. This observation suggests that some audio-visual hallucinations may be corrected by redirecting the fused representation toward evidence that remains accessible within the model.

To address this failure, we propose \textbf{CaRe (Conflict-aware Residual Intervention)}, a training-free inference-time method for selectively correcting modality misrouting. As illustrated in Figure~\ref{fig:care-pipeline}, CaRe follows a two-stage pipeline for detecting potential misrouting and correcting the corresponding fused representation. First, a text-only modality probe identifies the modality most likely required by the question and selects the corresponding counterfactual modality-specific branch by withholding the competing modality. A counterfactual conflict gate then compares its prediction with the original joint prediction and activates correction only when the two exhibit a hard decision conflict. For gate-positive samples, CaRe uses counterfactual residuals to steer the fused representation toward selected modality-specific reference, with strength adapted to counterfactual confidence. Importantly, the correction is applied to the original audio-visual representation while retaining both modalities. Experiments on CMM and AVHBench across two OLLMs show that CaRe consistently outperforms the evaluated inference-time hallucination-mitigation baselines. The main contributions of this work are summarized as follows:

\begin{itemize}
    \item We identify and characterize \emph{modality misrouting}, an underexplored yet recoverable failure mode of OLLMs in which question-relevant modality-specific evidence remains accessible but is not effectively reflected in the joint multimodal prediction. This perspective highlights a previously underexplored failure pattern in multimodal evidence integration that can give rise to cross-modal hallucinations.
    \item We propose CaRe, a training-free inference-time intervention framework that can be directly applied to existing open-source OLLMs without retraining or modifying model parameters. CaRe detects potential modality misrouting through text-only modality probing and counterfactual conflict gating, and corrects it by steering the fused representation with residual signals derived from the selected modality-specific counterfactual branch.
    \item We evaluate CaRe on two OLLMs and two audio-visual hallucination benchmarks. Results show that CaRe consistently outperforms strong inference-time baselines while preserving the complete audio-visual input in the multimodal inference path.
\end{itemize}

\section{Related Work}

\subsection{Omnimodal Large Language Models}

Recent multimodal foundation models have extended language models toward unified processing of audio, visual, and textual signals, including OneLLM \citep{han2024onellm}, VITA \citep{vita}, Qwen-Omni and MiniCPM-o \citep{Qwen2.5-Omni,minicpm,team2026qwen3,xu2025qwen3,minicpm4.5}, and OmniVinci \citep{ye2026omnivinci}. Despite increasingly strong joint perception, broader modality coverage does not necessarily ensure reliable evidence integration, as competing or imbalanced modalities may prevent question-relevant information from being effectively reflected in the joint prediction \citep{park2025assessing,zhang2025robust}. Our work focuses on this inference-time failure and uses recoverable modality-specific evidence to correct erroneous multimodal predictions without modifying the underlying OLLM.

\subsection{Hallucination Mitigation for Omnimodal Models}

Cross-modal hallucination remains a major reliability challenge for multimodal and omnimodal models \citep{zechen2024hallucination,OmniHallu}, as highlighted by benchmarks such as CMM \citep{cmm} and AVHBench \citep{sung2025avhbench}. Existing mitigation methods mainly rely on training-time alignment \citep{chen2026omnidpo,chaubey2026mod,yu2024rlhf,baid2026don,pan2026ground,jiang2024hallucination}, representation intervention \citep{li2023inference,chen2025ict,caa,su2025activation,li2025hidden,wu2025sharp,yin2026dynamic}, or training-free decoding correction \citep{vcd,jung2026avcd,chung2026mad,favero2024multi,huang2024opera,ma2026understanding}, including contrastive decoding, modality reweighting, and activation steering. These approaches primarily suppress hallucinated outputs or recalibrate modality contributions, while the recoverable setting where question-relevant unimodal evidence remains available but is not effectively reflected in the joint prediction has received less attention. CaRe targets this setting by detecting counterfactual disagreement and selectively correcting the joint representation while retaining the original audio-visual inference path.

\section{Motivation}

\begin{wrapfigure}{r}{0.4\linewidth}
    \centering
    \vspace{-1em}
    \includegraphics[width=\linewidth]{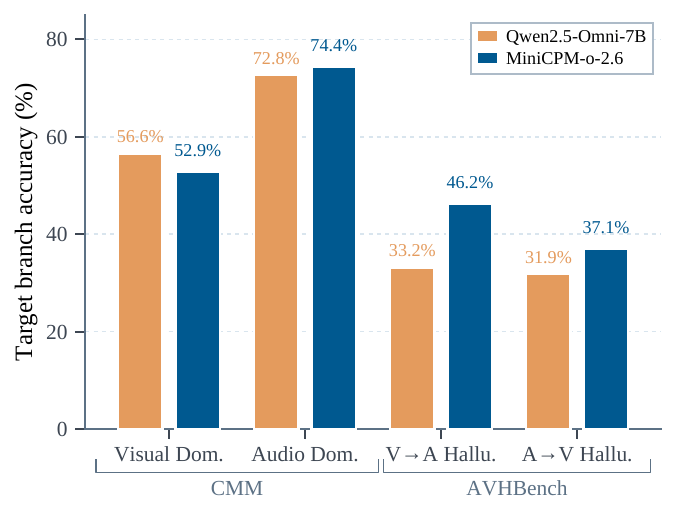}
    \vspace{-1.5em}
    \caption{Target-branch recovery among incorrect joint predictions
    on CMM and AVHBench (\%).}
    \label{fig:misrouting-prevalence}
    \vspace{-1.5em}
\end{wrapfigure}

Audio-visual hallucinations do not always coincide with a failure of the question-relevant unimodal view. In some cases, the joint audio-visual branch produces an incorrect prediction, while the branch retaining only the modality relevant to the question recovers the correct answer. We refer to this recoverable discrepancy between the joint and modality-specific predictions as modality misrouting. This observation suggests that some multimodal failures may still retain useful modality-specific evidence that can be exploited at inference time.

To quantify how often this recoverable pattern occurs, we analyze the Visual and Audio Dominance subsets of CMM together with the two cross-modal hallucination tasks of AVHBench. For each incorrect joint audio-visual prediction, we withhold the competing modality and test whether the question-relevant unimodal branch recovers the ground-truth answer. As shown in Figure~\ref{fig:misrouting-prevalence}, the recovery rate ranges from 52.9\% to 74.4\% on CMM and from 31.9\% to 46.2\% on AVHBench. The consistent recovery across both backbones, benchmarks, and interference directions shows that a nontrivial fraction of joint prediction failures remain recoverable from modality-specific evidence, motivating targeted correction of the joint representation.

\section{Preliminaries}
\label{sec:preliminaries}

\paragraph{Input configurations.}
Given an audio input $a$, a visual input $v$, and a textual question $q$, we consider three input configurations:
\[
x^{AV}=(a,v,q),\qquad
x^{A}=(a,\varnothing_V,q),\qquad
x^{V}=(\varnothing_A,v,q),
\]
where $\varnothing_A$ and $\varnothing_V$ denote withheld audio and visual inputs, respectively. Running the same frozen OLLM under these configurations induces a joint audio-visual branch $\mathcal{B}_{AV}$, an audio-only branch $\mathcal{B}_{A}$, and a video-only branch $\mathcal{B}_{V}$. We regard the two unimodal branches as counterfactual views of the original audio-visual input, obtained by withholding the competing modality.

\paragraph{Task-level decisions.}
Let $\mathcal{C}$ denote the finite answer space defined by the task. For binary questions, $\mathcal{C}=\{\texttt{Yes},\texttt{No}\}$; for multiple-choice questions, $\mathcal{C}$ consists of the provided answer options. For each branch $b\in\{AV,A,V\}$, let $z_c^b$ denote the model score associated with candidate $c\in\mathcal{C}$. We define the branch prediction as
\begin{equation}
\hat{y}_b=\arg\max_{c\in\mathcal{C}} z_c^b,
\end{equation}
and its decision margin as
\begin{equation}
m_b=z_{\hat{y}_b}^{b}-\max_{c\in\mathcal{C}\setminus\{\hat{y}_b\}} z_c^b.
\end{equation}
A larger $m_b$ indicates a more decisive prediction, while $m_b=0$ indicates that the branch does not have a unique preferred answer. In our evaluations, $z_c^b$ is the next-token logit assigned to the candidate answer label under the evaluation prompt.

\paragraph{Question-level modality routing.}
Inspired by the modality self-assessment mechanism of MAD \citep{chung2026mad},
we use the OLLM itself to estimate which sensory modality is required by the question. We adopt a text-only routing probe: both audio and visual inputs are withheld, and the original question is followed by the prompt: "\textit{To answer this question, which modality is needed (audio, video, or both):}"
Let $\rho_A$, $\rho_V$, and $\rho_B$ denote the logits assigned to \texttt{audio}, \texttt{video}, and \texttt{both}, respectively, and define:
\begin{equation}
(p_a,p_v,p_b) =\operatorname{softmax}(\rho_A,\rho_V,\rho_B).
\end{equation}
Because the probe observes only the question, it provides a question-level modality-demand signal without accessing the audio-visual content of the current sample. The probe is used only to select a candidate counterfactual reference; it is not assumed to identify the factually correct modality.
We select a modality-specific counterfactual reference only when the probe uniquely favors
audio or video:
\begin{equation}
r=
\begin{cases}
A, & p_a>\max(p_v,p_b),\\
V, & p_v>\max(p_a,p_b),\\
\varnothing, & \text{otherwise}.
\end{cases}
\end{equation}
Samples routed to \texttt{both}, or for which no unique maximum exists, are assigned $r=\varnothing$. Thus, the routing signal only determines which unimodal branch, if any, serves as a candidate reference; it does not itself trigger an intervention. We evaluate the robustness of this routing signal to prompt formulation and option ordering in Section~\ref{sec:ablation}.

\section{Method}
\label{sec:method}

\subsection{Counterfactual Conflict Gate}
\label{sec:counterfactual-conflict-gate}

The routing signal identifies a candidate modality-specific reference, but disagreement with the joint audio-visual prediction is required before CaRe applies any correction. Given $r\in\{A,V,\varnothing\}$, we define the counterfactual conflict gate as
\begin{equation}
g=
\mathbb{I}\left[
r\in\{A,V\} \;\land\;
m_r>0 \;\land\;
m_{AV}>0 \;\land\;
\hat{y}_r\neq\hat{y}_{AV}
\right].
\end{equation}

Thus, $g=1$ only when the routing probe provides a valid unimodal reference and both the counterfactual and joint branches make unambiguous but conflicting decisions. We refer to this disagreement as a \emph{hard counterfactual conflict}. All other samples receive $g=0$ and retain the original joint prediction. When $g=1$, the selected branch $r$ is used as the counterfactual reference for the residual intervention described in Section~\ref{sec:intervention}.

\subsection{Adaptive Residual Intervention}
\label{sec:intervention}

The counterfactual conflict gate identifies samples for which the joint prediction disagrees with the selected modality-specific counterfactual reference, but it does not specify how the joint representation should be corrected without discarding complementary multimodal context. We therefore use the selected counterfactual branch as a representation-level reference and steer the original audio-visual inference path toward it. 

For a gate-positive sample, let $h_l^r\in\mathbb{R}^d$ denote the post-residual hidden state at the final prompt-token position of decoder layer $l$ from the selected counterfactual branch $r$, and let $h_l^{AV}\in\mathbb{R}^d$ denote the corresponding state on the audio-visual intervention path. We define the layer-wise counterfactual residual as
\begin{equation}
\Delta h_l^r
=
h_l^r-h_l^{AV}.
\end{equation}
This residual gives a sample-specific direction from the current audio-visual representation toward its selected counterfactual reference. Importantly, $h_l^{AV}$ is the current state during the intervention forward pass and therefore already reflects corrections made at preceding layers. The residual at each layer is consequently computed relative to the actual evolving audio-visual representation rather than the original unmodified state. CaRe uses this residual only to adjust the original audio-visual representation, while both modalities remain available throughout subsequent computation. 

\subsubsection{Confidence-Adaptive Intervention Strength}

A counterfactual conflict alone does not indicate how strongly the candidate branch should influence the joint representation. In particular, a counterfactual prediction that barely separates its top candidates provides a weaker correction signal than one that makes a decisive prediction. We therefore adapt the intervention strength to the decision confidence of the selected branch.

We convert its decision margin $m_r$ into a bounded confidence score:
\begin{equation}
u_r=
2\sigma(m_r)-1
\in (0,1),
\end{equation}
where $\sigma(\cdot)$ denotes the sigmoid function. Larger values of $u_r$ correspond to greater separation between the selected answer and its strongest alternative. Given a confidence threshold $\eta$, CaRe selects either a weak or strong intervention range:
\begin{equation}
[\alpha_{\min},\alpha_{\max}]
=
\begin{cases}
[\alpha_{\min}^{\mathrm{W}},\alpha_{\max}^{\mathrm{W}}],
& u_r<\eta,\\
[\alpha_{\min}^{\mathrm{S}},\alpha_{\max}^{\mathrm{S}}],
& u_r\geq\eta.
\end{cases}
\end{equation}
Low-confidence counterfactual predictions therefore receive conservative corrections, whereas more decisive predictions permit stronger intervention. 

\subsubsection{Depth-Decayed Residual Intervention}

Applying repeated corrections with the same strength across multiple decoder layers can lead to excessive perturbation, particularly near the output. We therefore distribute the intervention over a predefined set of decoder layers while gradually reducing its strength with depth, allowing earlier corrections to propagate through subsequent computation and later layers to make milder refinements. Let
\[
\mathcal{L}_{\mathrm{int}}
=
\{l_1,\ldots,l_K\},
\qquad
l_1<\cdots<l_K,
\quad K\geq 2,
\]
denote the intervention layers. Given the confidence-dependent range
$[\alpha_{\min},\alpha_{\max}]$, the coefficient at layer $l_j$ is
\begin{equation}
\alpha_j
=
\alpha_{\min}
+
(\alpha_{\max}-\alpha_{\min})
\left(
\frac{K-j}{K-1}
\right)^{\gamma},
\qquad
j=1,\ldots,K,
\end{equation}
where $\gamma>0$ controls the decay rate. Thus, $\alpha_j$ decreases monotonically from $\alpha_{\max}$ at $l_1$ to $\alpha_{\min}$ at $l_K$. Following prior activation-steering work that uses the last input-token representation as a compact site for controlling subsequent generation \citep{stolfo2025improving}, CaRe modifies only the final prompt-token state during prefilling. At each intervention layer $l_j$, we apply
\begin{equation}
\tilde{h}_{l_j}^{AV}
=
h_{l_j}^{AV}
+
\alpha_j\Delta h_{l_j}^r
=
h_{l_j}^{AV}
+
\alpha_j
\left(
h_{l_j}^r-h_{l_j}^{AV}
\right).
\end{equation}
The corrected state $\tilde{h}_{l_j}^{AV}$ is propagated through the subsequent decoder layers, so later interventions operate on the updated audio-visual representation. All other prompt-token positions remain unchanged. After prefilling, generation proceeds autoregressively from the resulting audio-visual state without further intervention on newly generated tokens.

\section{Experiments}

\subsection{Experimental Setup}
\label{subsec:setup}

\paragraph{Models and Benchmarks.}
We evaluate CaRe on Qwen2.5-Omni-3B/7B \citep{Qwen2.5-Omni} and MiniCPM-o-2.6 \citep{minicpm}, two omnimodal large language models that jointly process audio, visual, and textual inputs. We use two benchmarks for cross-modal hallucination evaluation: CMM \cite{cmm} and AVHBench \citep{sung2025avhbench}. CMM examines hallucinations arising from modality over-reliance; we focus on its Visual Dominance and Audio Dominance subsets. AVHBench evaluates whether information from one modality induces false perceptions in the other; we use its Video-driven Audio Hallucination and Audio-driven Video Hallucination tasks. 

\paragraph{Baselines and Metrics.}
We compare CaRe with the unmodified backbone and four inference-time hallucination-mitigation baselines: Visual Contrastive Decoding (VCD) \citep{vcd}, Image-Object Cross-Level Trusted Intervention (ICT) \citep{chen2025ict}, Audio-Visual Contrastive Decoding (AVCD) \citep{jung2026avcd}, and Modality-Adaptive Decoding (MAD) \citep{chung2026mad}. These methods respectively represent visual contrastive decoding, internal representation intervention, audio-visual contrastive
decoding, and modality-adaptive decoding. Further details of the baseline methods are provided in Appendix~\ref{app:implementation}. We report accuracy (Acc.) and F1 score, with \texttt{Yes} treated as the positive class. \emph{Overall Acc.} is computed by pooling samples across the two reported subsets or tasks within each benchmark. All values are reported in percentage.

\paragraph{Implementation Details.}

Qwen2.5-Omni-7B and MiniCPM-o-2.6 each contain 28 language-decoder layers, while Qwen2.5-Omni-3B contains 36 layers. We apply CaRe to the final three decoder layers, corresponding to layers 25--27 for Qwen2.5-Omni-7B and MiniCPM-o-2.6, and layers 33--35 for Qwen2.5-Omni-3B. We use greedy decoding with temperature 0 for deterministic evaluation. Implementation details, hyperparameter configurations, and sensitivity analyses are provided in Appendix~\ref{app:implementation}.

\begin{table*}[t]
\vspace{-0.5em}
\centering
\caption{\textbf{Main results on CMM and AVHBench (\%).} We evaluate
Qwen2.5-Omni-7B and MiniCPM-o-2.6 on the Visual Dominance and Audio Dominance subsets of CMM, and on the Video-driven Audio Hallucination and Audio-driven Video Hallucination tasks of AVHBench. We report Accuracy (Acc.) and F1, with \texttt{Yes} treated as the positive class. \emph{Overall Acc.} is computed over all samples in the two reported subsets or tasks. Best results are highlighted in bold.}
\vspace{-0.5em}
\label{tab:main-results}
\small
\setlength{\tabcolsep}{3.5pt}
\renewcommand{\arraystretch}{1.12}
\begin{tabular}{@{}ll cc cc c cc cc c@{}}
\toprule
& & \multicolumn{5}{c}{\textbf{CMM}}
  & \multicolumn{5}{c}{\textbf{AVHBench}} \\
\cmidrule(lr){3-7}\cmidrule(lr){8-12}
\textbf{Backbone} & \textbf{Method}
& \multicolumn{2}{c}{\shortstack[c]{Visual\\Dominance}}
& \multicolumn{2}{c}{\shortstack[c]{Audio\\Dominance}}
& \multirow{2}{*}{\shortstack[c]{Overall\\Acc.}}
& \multicolumn{2}{c}{\shortstack[c]{Video-Driven\\Audio Hall.}}
& \multicolumn{2}{c}{\shortstack[c]{Audio-Driven\\Video Hall.}}
& \multirow{2}{*}{\shortstack[c]{Overall\\Acc.}} \\

\cmidrule(lr){3-4}
\cmidrule(lr){5-6}
\cmidrule(lr){8-9}
\cmidrule(lr){10-11}

& & Acc. & F1 
& Acc. & F1 
&
& Acc. & F1 
& Acc. & F1
& \\
\midrule

\multirow{6}{*}{\shortstack[l]{Qwen2.5-\\Omni-7B}}
& Baseline
& 69.3 & 75.7 & 71.8 & 77.0 & 70.5
& 67.6 & 75.3 & 74.1 & 77.4 & 69.8 \\

& +VCD
& 75.5 & 79.4 & 66.5 & 73.5 & 71.0
& 68.1 & 75.4 & 77.4 & 78.0 & 71.2 \\

& +ICT 
& 75.3 & 79.1 & 66.5 & 73.6 & 70.9
& 68.8 & 75.6 & 76.7 & 78.4 & 71.4 \\

& +AVCD 
& 67.8 & 74.9 & 70.5 & 76.2 & 69.1
& 77.9 & 81.4 & 81.7 & 82.8 & 79.2 \\

& +MAD 
& 69.5 & 75.9 & 75.5 & 79.2 & 72.5
& 79.0 & 82.1 & 85.0 & 84.5 & 81.0\\

& \carecell{\textbf{+CaRe}}
& \carecell{\textbf{82.0}} & \carecell{\textbf{84.1}}
& \carecell{\textbf{89.8}} & \carecell{\textbf{89.8}}
& \carecell{\textbf{85.9}}
& \carecell{\textbf{80.9}} & \carecell{\textbf{83.2}}
& \carecell{\textbf{85.1}} & \carecell{\textbf{84.5}}
& \carecell{\textbf{82.3}} \\

\midrule

\multirow{6}{*}{\shortstack[l]{MiniCPM-\\o-2.6}}
& Baseline
& 78.3 & 81.2 & 67.8 & 73.6 & 73.0
& 72.8 & 74.4 & 74.7 & 75.7 & 73.4 \\

& +VCD
& 72.3 & 76.3 & 61.0 & 69.5 & 66.6
& 73.9 & 75.0 & 74.0 & 74.5 & 73.9 \\

& +ICT 
& 72.3 & 76.1 & 62.8 & 70.8 & 67.5
& 72.3 & 73.8 & 75.9 & 76.4 & 73.5 \\

& +AVCD 
& 77.8 & 80.9 & 67.5 & 73.5 & 72.6
& 78.9 & 80.6 & 74.1 & 75.9 & 77.3  \\

& +MAD 
& 81.8 & 83.8 & 72.8 & 76.0 & 77.3
& 79.8 & 81.9 & 73.6 & 74.6 & 77.8\\

& \carecell{\textbf{+CaRe}}
& \carecell{\textbf{85.3}} & \carecell{\textbf{86.4}}
& \carecell{\textbf{88.5}} & \carecell{\textbf{88.3}}
& \carecell{\textbf{86.9}}
& \carecell{\textbf{81.7}} & \carecell{\textbf{83.3}}
& \carecell{\textbf{77.1}} & \carecell{\textbf{77.4}}
& \carecell{\textbf{80.2}} \\

\bottomrule
\end{tabular}
\vspace{-0.5em}
\end{table*}

\subsection{Main Results}

Table~\ref{tab:main-results} reports the main results on CMM and AVHBench.
CaRe achieves the best overall accuracy for both backbones on both benchmarks, and obtains the highest score in every reported subset and task. These consistent gains over contrastive decoding, representation intervention, and modality-adaptive decoding baselines demonstrate that CaRe effectively mitigates cross-modal hallucinations across different model architectures and failure settings.

\paragraph{Results on CMM.}
CaRe substantially improves robustness against modality over-reliance. On Qwen2.5-Omni-7B, CaRe reaches an overall accuracy of 85.9\%, exceeding the strongest baseline, MAD, by 13.4 points. On MiniCPM-o-2.6, CaRe achieves 86.9\%, outperforming MAD by 9.6 points. The gains are particularly pronounced on the Audio Dominance subset: CaRe improves accuracy from 75.5\% to 89.8\% on Qwen2.5-Omni-7B and from 72.8\% to 88.5\% on MiniCPM-o-2.6, relative to the strongest baseline. CaRe also consistently improves F1 scores on both dominance subsets, indicating more reliable decisions under modality-conflict scenarios.

\begin{wraptable}{r}{0.46\columnwidth}
\vspace{-0.6em}
\setlength{\abovecaptionskip}{3pt}
\setlength{\belowcaptionskip}{-6pt}
\setlength{\tabcolsep}{2.3pt}
\renewcommand{\arraystretch}{1.05}

\centering
\caption{Scaling robustness on Qwen2.5-Omni models. Gains over Base are shown in parentheses.}
\label{tab:scaling}

\small
\begin{tabular}{llcc}
\toprule
\multirow{2}{*}{\textbf{Backbone}} & \multirow{2}{*}{\textbf{Method}} & \textbf{CMM} & \textbf{AVHBench} \\
\cmidrule(lr){3-3} \cmidrule(lr){4-4}
 & & Acc. & Acc. \\
\midrule

\multirow{2}{*}{\makecell[l]{Qwen2.5-\\Omni-3B}}
 & Base  & 76.9 & 76.4 \\
 & \textbf{+CaRe} & \textbf{82.5} \gain{5.6} & \textbf{81.0} \gain{4.6} \\

\addlinespace[3pt]
\cmidrule(lr){1-4} 
\addlinespace[3pt]

\multirow{2}{*}{\makecell[l]{Qwen2.5-\\Omni-7B}}
 & Base  & 70.5 & 69.8 \\
 & \textbf{+CaRe} & \textbf{85.9} \gain{15.4} & \textbf{82.3} \gain{12.5} \\

\bottomrule
\end{tabular}

\vspace{-1.0em}
\end{wraptable}

\paragraph{Results on AVHBench.}
CaRe also achieves the best overall accuracy on both cross-modal hallucination tasks in AVHBench. For Qwen2.5-Omni-7B, CaRe attains 82.3\% overall accuracy, surpassing MAD by 1.3 points. For MiniCPM-o-2.6, it reaches 80.2\%, improving over the strongest baseline, MAD, by 2.4 points. CaRe obtains the highest accuracy and F1 score on both Video-driven Audio Hallucination and Audio-driven Video Hallucination tasks for both backbones. These results show that the proposed conflict-aware residual intervention generalizes to hallucinations caused by interference in either modality direction.

\paragraph{Robustness across model scales.} To examine whether CaRe depends on model capacity, we additionally evaluate Qwen2.5-Omni-3B, providing a controlled comparison with its 7B counterpart within the same model family. On the 3B model, CaRe improves overall accuracy from 76.9\% to 82.5\% on CMM and from 76.4\% to 81.0\% on AVHBench, yielding gains of 5.6 and 4.6 points, respectively. Together with the improvements observed on Qwen2.5-Omni-7B, these results show that CaRe remains effective across model scales rather than being specific to a single capacity setting. Detailed category-wise results are provided in Appendix~\ref{app:scaling}.

\subsection{Ablation Studies}
\label{sec:ablation}

We conduct component-wise ablations to examine the main design choices of CaRe. Unless otherwise specified, all experiments are conducted on CMM and aggregate results over the Visual and Audio Dominance subsets. Each ablation varies one component while keeping the remaining settings fixed.

\begin{figure*}[t]
    \vspace{-0.5em}
    \centering
    \captionsetup{font=small,justification=raggedright,singlelinecheck=false}
    \begin{minipage}[t]{0.32\textwidth}
        \centering
        \includegraphics[width=\linewidth]{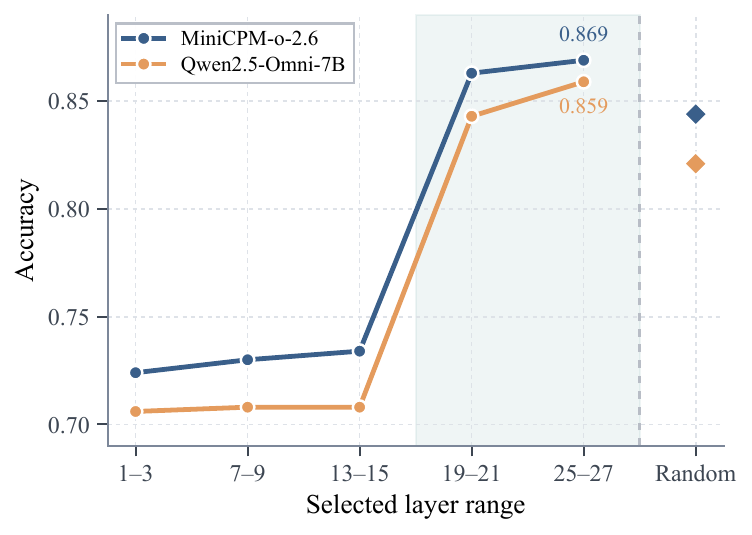}
        \vspace{-1em}
        \caption{\textbf{Intervention-layer location.} Late layers 25--27 achieve the best accuracy across both backbones.}
        \label{fig:layer-ablation}
    \end{minipage}
    \hfill
    \begin{minipage}[t]{0.32\textwidth}
        \centering
        \includegraphics[width=\linewidth]{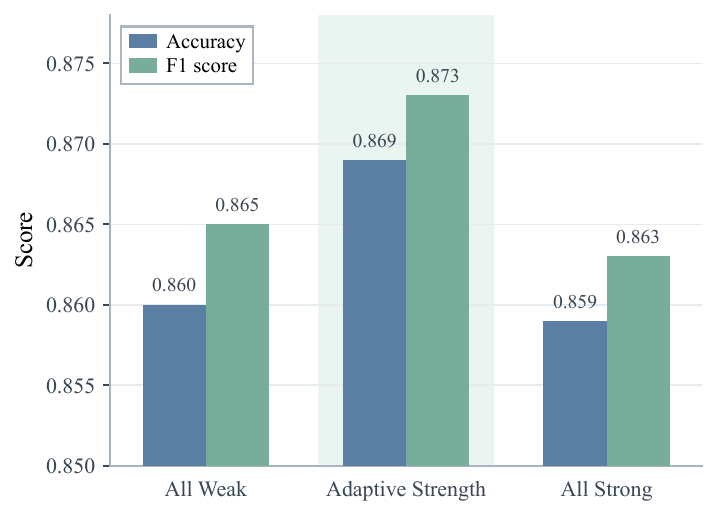}
        \vspace{-1em}
        \caption{\textbf{Confidence-adaptive strength.} Adaptive strength outperforms uniformly weak and strong interventions.}
        \label{fig:strength-ablation}
    \end{minipage}
    \hfill
    \begin{minipage}[t]{0.32\textwidth}
        \centering
        \includegraphics[width=\linewidth]{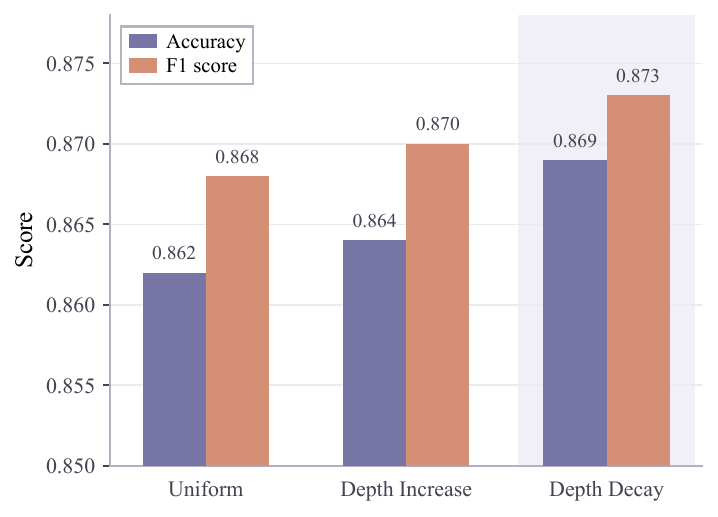}
        \vspace{-1em}
        \caption{\textbf{Depth-decayed intervention.} Depth decay outperforms uniform and depth-increasing schedules.}
        \label{fig:depth-decay-ablation}
    \end{minipage}
\end{figure*}

\paragraph{Residual intervention vs.\ branch selection.}

To isolate the benefit of representation-level correction, we compare CaRe with CF-SELECT, which directly outputs the counterfactual prediction $\hat{y}^r$ when a conflict is detected ($g=1$).
Since both methods behave identically on gate-negative samples, we evaluate directly on the gate-positive subset where their decisions diverge.
As shown in Table~\ref{tab:gate_positive_ablation}, CaRe consistently outperforms CF-SELECT across both backbones and both hallucination directions on AVHBench, achieving accuracy gains of up to 10.82 points.
This confirms that the advantage of CaRe stems from steering the multimodal representation rather than trivially switching to a unimodal branch: by using the counterfactual signal as a continuous residual direction, CaRe resolves modality misrouting while retaining complementary cross-modal context.

\begin{table*}[t]
\centering
\small
\caption{\textbf{Residual intervention vs. direct branch selection on conflict-gated samples from AVHBench (\%).} 
Evaluations are conducted exclusively on conflict-gated samples ($g=1$). 
CF-SELECT directly outputs the counterfactual unimodal prediction $\hat{y}^r$, whereas CaRe applies continuous residual steering to the joint representation. 
Accuracy gains of CaRe over CF-SELECT are reported in parentheses at the bottom right.}
\vspace{-0.5em}
\label{tab:gate_positive_ablation}
\begin{tabular}{llcc}
\toprule
\textbf{Backbone} & \textbf{Method} & \textbf{Video-Driven Audio Hall.} & \textbf{Audio-Driven Video Hall.} \\
\midrule
\multirow{2}{*}{Qwen2.5-Omni-7B}
 & CF-SELECT & 54.22 & 53.61 \\
 & \textbf{+CaRe} & \textbf{63.05}$_{\,\textcolor{gainblue}{(+8.83)}}$ & \textbf{62.89}$_{\,\textcolor{gainblue}{(+9.28)}}$ \\
\midrule
\multirow{2}{*}{MiniCPM-o-2.6} 
 & CF-SELECT & 48.56 & 60.38 \\
 & \textbf{+CaRe} & \textbf{59.38}$_{\,\textcolor{gainblue}{(+10.82)}}$ & \textbf{64.15}$_{\,\textcolor{gainblue}{(+3.77)}}$ \\
\bottomrule
\end{tabular}
\end{table*}

\begin{figure*}[t]
    \centering
    \includegraphics[width=0.8\textwidth]{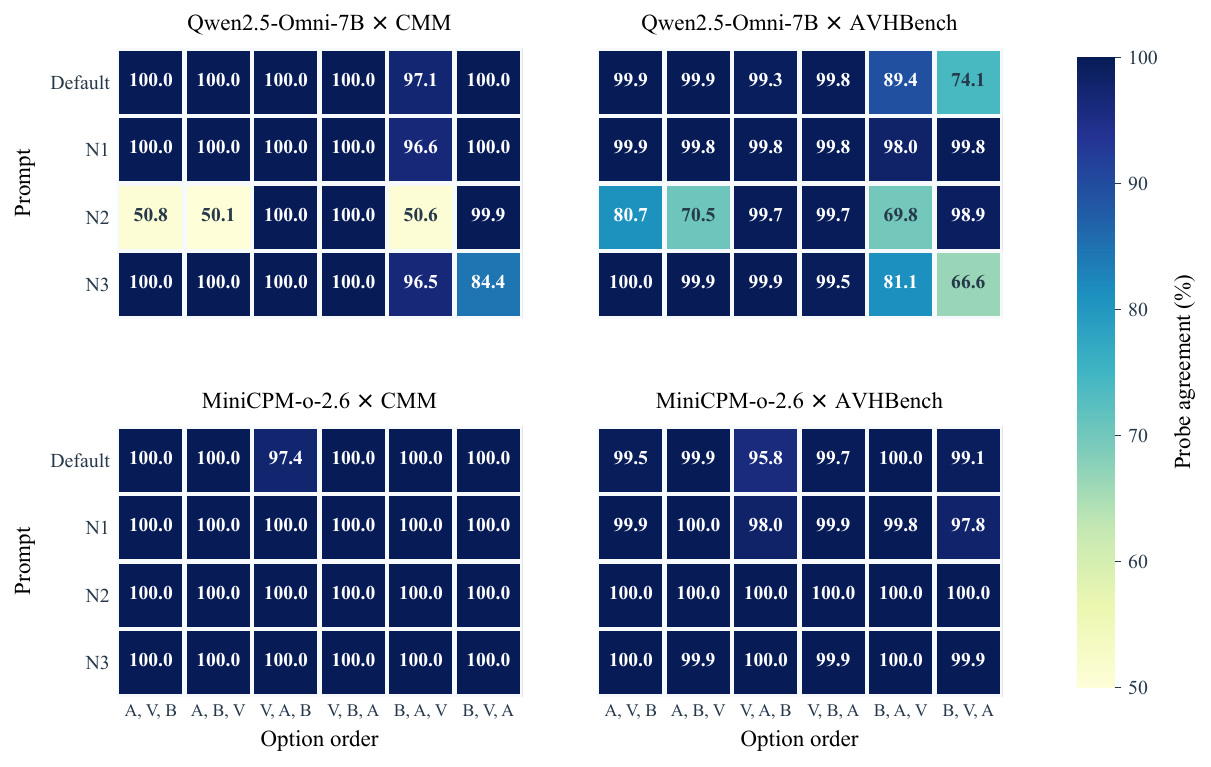}
    \vspace{-1em}
    \caption{\textbf{Robustness of text-only modality routing.} Probe agreement (\%) with the task-defined target modality across prompt formulations and all six modality-label orderings. Results are shown for both backbones on CMM and AVHBench; prompt templates are provided in Appendix~\ref{app:probe-prompts}.}
    \label{fig:probe-robustness}
\end{figure*}

\paragraph{Robustness of text-only modality routing.}
Since CaRe relies on the text-only probe to select a counterfactual reference, we examine the robustness of its routing decisions to prompt formulation and option ordering. For CMM and AVHBench, the question-relevant modality is specified by task construction, allowing us to measure probe agreement with the task-defined target modality. As shown in Figure~\ref{fig:probe-robustness}, we evaluate four prompt formulations and all six permutations of \texttt{audio}, \texttt{video}, and \texttt{both} across the four backbone--benchmark settings. Among the resulting 96 configurations, 85 achieve at least 95\% agreement, while the default configuration used in all main experiments achieves 99.5--100\% agreement across the four settings. Overall, the routing signal is broadly stable, with some model-specific sensitivity to prompt and option ordering. The exact probe formulations are provided in Appendix~\ref{app:probe-prompts}.

\paragraph{Intervention-layer location.}
Figure~\ref{fig:layer-ablation} compares different three-layer intervention
ranges. Interventions in early and intermediate layers provide limited
benefits, whereas performance improves sharply in later layers. Compared with layers 19--21, the default range of layers 25--27 further improves accuracy by 1.6 points on Qwen2.5-Omni-7B and 0.6 points on MiniCPM-o-2.6. It also outperforms random layer selection by 3.8 and 2.5 points, respectively. These results support applying residual correction near the output stage, where the fused representation is more directly associated with the final decision.

\paragraph{Confidence-adaptive intervention strength.}
Figure~\ref{fig:strength-ablation} compares adaptive strength with uniformly weak and uniformly strong interventions on CMM using MiniCPM-o-2.6. Adaptive strength improves accuracy and F1 by 0.9 and 0.8 points over the all-weak setting, and by 1.0 point on both metrics over the all-strong setting. This suggests that weak correction can be insufficient for decisive counterfactual predictions, whereas uniformly strong correction may over-perturb less certain cases. Adapting the intervention strength to counterfactual confidence provides a better balance.

\paragraph{Depth-decayed residual intervention.}
Figure~\ref{fig:depth-decay-ablation} compares the proposed depth-decayed schedule with uniform strength and depth-increasing strength on CMM using MiniCPM-o-2.6. The depth-decayed schedule performs best, achieving 0.869 accuracy and 0.873 F1, compared with 0.862/0.868 for uniform strength and 0.864/0.870 for depth-increasing strength. The gain does not arise merely from varying the intervention coefficients across layers; assigning stronger corrections to earlier layers in the intervention range and weaker corrections to deeper layers is particularly effective.

\subsection{Performance on General Audio-Visual QA Tasks}

To assess CaRe beyond hallucination-focused benchmarks, we evaluate it on Daily-Omni \citep{zhou2025daily}, a general audio-visual QA benchmark covering event sequencing, audio-visual alignment, inference, reasoning, contextual understanding, and comparative analysis. As shown in Table~\ref{tab:daily-omni}, CaRe increases overall accuracy from 61.5\% to 61.8\% on Qwen2.5-Omni-7B and from 61.0\% to 61.5\% on MiniCPM-o-2.6. The most consistent category-level gains are observed in inference and reasoning, with improvements of 1.3 points on inference for both backbones and 1.2/1.1 points on reasoning. CaRe also yields slightly higher accuracy on both 30-second and 60-second videos. Overall, these results indicate that applying CaRe does not compromise aggregate performance on general audio-visual QA, while yielding small overall accuracy gains on both backbones.

\begin{table*}[t]
\vspace{-0.5em}
\centering
\caption{\textbf{Accuracy on Daily-Omni (\%).}
Results are reported across question categories and video durations.
The better result within each backbone is highlighted in bold.}
\vspace{-0.5em}
\label{tab:daily-omni}
\scriptsize
\setlength{\tabcolsep}{3.4pt}
\renewcommand{\arraystretch}{1.12}
\begin{tabular}{llccccccccc}
\toprule
\multirow[c]{3}{*}{\textbf{Backbone}}
& \multirow[c]{3}{*}{\textbf{Method}}
& \multirow[c]{3}{*}{\textbf{All}}
& \multicolumn{6}{c}{\textbf{Question Category}}
& \multicolumn{2}{c}{\textbf{Duration}} \\
\cmidrule(lr){4-9}
\cmidrule(lr){10-11}
&
&
& \makecell[c]{\textbf{Event}\\\textbf{Seq.}}
& \makecell[c]{\textbf{AV Event}\\\textbf{Align.}}
& \makecell[c]{\textbf{Inference}}
& \makecell[c]{\textbf{Reasoning}}
& \makecell[c]{\textbf{Context}\\\textbf{Understanding}}
& \makecell[c]{\textbf{Comparative}}
& \makecell[c]{\textbf{30s}}
& \makecell[c]{\textbf{60s}} \\
\midrule

\multirow{2}{*}{Qwen2.5-Omni-7B}
& Base
& 61.5
& 55.6
& 50.4
& 75.3
& 75.4
& \textbf{57.5}
& \textbf{66.4}
& 62.6
& 60.2 \\

& CaRe
& \textbf{61.8}
& \textbf{56.2}
& \textbf{51.3}
& \textbf{76.6}
& \textbf{76.6}
& 56.5
& 64.9
& \textbf{62.9}
& \textbf{60.5} \\

\midrule

\multirow{2}{*}{MiniCPM-o-2.6}
& Base
& 61.0
& 55.6
& \textbf{48.7}
& 76.0
& 72.6
& \textbf{60.6}
& 63.4
& 61.1
& 60.9 \\

& CaRe
& \textbf{61.5}
& \textbf{55.9}
& \textbf{48.7}
& \textbf{77.3}
& \textbf{73.7}
& \textbf{60.6}
& \textbf{64.1}
& \textbf{61.8}
& \textbf{61.1} \\

\bottomrule
\end{tabular}
\end{table*}

\section{Conclusion}

Our results show that some audio-visual hallucinations occur even when task-relevant evidence remains recoverable from a modality-specific view, motivating modality misrouting as a useful characterization of such failures. Based on this observation, we introduce CaRe, which detects conflicts between the joint prediction and a question-relevant counterfactual view, and uses the resulting residual signal to correct fused representations. Its confidence-adaptive and depth-decayed intervention enables targeted correction at middle-to-late decoder layers without updating model parameters. Experiments on CMM and AVHBench demonstrate substantial improvements across two OLLMs, while Daily-Omni results show that CaRe also preserves and modestly improves general audio-visual question-answering performance.

\paragraph{Limitations and Future Work.}
CaRe currently assumes that useful task-relevant evidence remains accessible in a modality-specific branch, and its conflict gate is formulated over a discrete task-defined answer space where branch predictions can be directly compared. As a result, it may be less effective when relevant evidence is weakly encoded, when modality misrouting does not manifest as an explicit prediction conflict, or when the task does not provide a readily scorable candidate answer space. In addition, the text-only routing probe and counterfactual branches introduce extra inference overhead. Future work may explore softer conflict signals, more general routing criteria for open-ended generation, and more efficient single-pass implementations.

\subsection*{AI use statement}

Generative AI tools were used to assist with language editing, improving the clarity and readability of the manuscript, and organizing and formatting data. They were not used to formulate the research methodology, generate or interpret experimental results, or derive the scientific conclusions of this work. All AI-assisted content was reviewed and verified by the authors, who take full responsibility for the final content of the paper.

\subsection*{Reproducibility Statement}

We provide detailed information to facilitate reproduction of our results. Sections~\ref{sec:counterfactual-conflict-gate}--\ref{sec:intervention} describe the complete formulation of CaRe, and Section~\ref{subsec:setup} specifies the evaluated models, benchmarks, metrics, and experimental protocol. Appendix~\ref{app:sensitivity} provides additional ablations and
hyperparameter analyses, including the counterfactual conflict gate, intervention strength, confidence threshold, and their joint sensitivity. Appendix~\ref{app:implementation} provides the complete CaRe inference algorithm, hyperparameter configurations, selection protocol, and implementation details for all baselines. Appendix~\ref{app:probe-prompts} reports the exact prompt formulations used for the text-only modality probe, while Appendix~\ref{app:dataset_details} provides additional details on the evaluation datasets. Finally, Appendix~\ref{app:computational} reports end-to-end inference cost,
runtime decomposition, and the potential latency reduction under ideal branch-level parallel execution. 

\bibliography{arxiv}
\bibliographystyle{arxiv}

\appendix
\clearpage

\section*{Appendix Overview}

\begin{table}[h]
\centering
\renewcommand{\arraystretch}{1.15}
\setlength{\tabcolsep}{8pt}

\begin{tabularx}{\linewidth}{
    |>{\raggedright\arraybackslash}p{0.28\linewidth}
    |>{\raggedright\arraybackslash}X|
}
\hline

\rowcolor{appendixheader}
\textbf{Appendix} & \textbf{Title} \\
\hline

\rowcolor{appendixpurple}
\textbf{Appendix~\ref{app:sensitivity}} &
\textbf{Additional Ablation and Hyperparameter Analysis} \\

\hspace{1.5em}$\hookrightarrow$ Appendix~\ref{app:gate} &
Ablation of the Counterfactual Conflict Gate \\

\hspace{1.5em}$\hookrightarrow$ Appendix~\ref{app:alpha-impact} &
Impact of the Intervention Strength Scale $\lambda_{\alpha}$ \\

\hspace{1.5em}$\hookrightarrow$ Appendix~\ref{app:eta-impact} &
Impact of the Confidence Threshold $\eta$ \\

\hspace{1.5em}$\hookrightarrow$ Appendix~\ref{app:joint-sensitivity} &
Joint Sensitivity to $\lambda_{\alpha}$ and $\eta$ \\
\hline

\rowcolor{appendixpurple}
\textbf{Appendix~\ref{app:scaling}} &
\textbf{Scaling Analysis} \\
\hline

\rowcolor{appendixpurple}
\textbf{Appendix~\ref{app:implementation}} &
\textbf{Reproducibility Details} \\

\hspace{1.5em}$\hookrightarrow$ Appendix~\ref{app:algorithm} &
CaRe Inference Algorithm \\

\hspace{1.5em}$\hookrightarrow$ Appendix~\ref{app:hyperparameters} &
Hyperparameter Configurations \\

\hspace{1.5em}$\hookrightarrow$ Appendix~\ref{app:baselines} &
Baseline Implementation Details \\
\hline

\rowcolor{appendixpurple}
\textbf{Appendix~\ref{app:probe-prompts}} &
\textbf{Prompt Formulations for the Text-Only Modality Probe} \\
\hline

\rowcolor{appendixpurple}
\textbf{Appendix~\ref{app:dataset_details}} &
\textbf{Additional Dataset Details} \\

\hspace{1.5em}$\hookrightarrow$ Appendix~\ref{app:cmm} &
CMM \\

\hspace{1.5em}$\hookrightarrow$ Appendix~\ref{app:avhbench} &
AVHBench \\

\hspace{1.5em}$\hookrightarrow$ Appendix~\ref{app:daily_omni} &
Daily-Omni \\
\hline

\rowcolor{appendixpurple}
\textbf{Appendix~\ref{app:computational}} &
\textbf{Computational Analysis} \\

\hline
\end{tabularx}
\end{table}

\section{Additional Ablation and Hyperparameter Analysis}
\label{app:sensitivity}

\subsection{Ablation of the Counterfactual Conflict Gate}
\label{app:gate}

We compare the full CaRe gate with a \emph{routing-only} variant that applies residual correction whenever the probe identifies a unique modality-specific candidate, without requiring a counterfactual conflict. As shown in Table~\ref{tab:gate-ablation}, removing the conflict condition increases intervention coverage by approximately 75 percentage points on both backbones, yet leaves accuracy unchanged. Moreover, none of these additional interventions changes prediction correctness. This result indicates that the conflict gate effectively excludes redundant corrections and concentrates intervention on samples with an explicit disagreement between the joint and counterfactual decisions.

\begin{table}[h]
\centering
\caption{\textbf{Ablation of the counterfactual conflict gate on CMM.}
\emph{Routing-only} applies residual correction whenever the probe returns a
unique unimodal candidate. \emph{Help/Harm} counts additional routing-only
interventions that change an originally incorrect/correct prediction into a
correct/incorrect one, respectively.}
\label{tab:gate-ablation}
\small
\setlength{\tabcolsep}{3pt}
\renewcommand{\arraystretch}{1.12}
\begin{tabular}{lcccc}
\toprule
\multicolumn{1}{c}{\multirow{2}{*}{\textbf{Backbone}}}
& \multicolumn{2}{c}{\textbf{Intervention Coverage}}
& \textbf{Accuracy}
& \multirow{2}{*}{\shortstack[c]{\textbf{Extra Interventions}\\\textbf{(Help / Harm)}}} \\
\cmidrule(lr){2-3}\cmidrule(lr){4-4}
& \textbf{Conflict-gated}
& \textbf{Routing-only}
& \textbf{Conflict / Routing}
& \\
\midrule
Qwen2.5-Omni-7B
& 167/800 (20.9\%)
& 773/800 (96.6\%)
& 85.9 / 85.9
& 606 (0 / 0) \\

MiniCPM-o-2.6
& 174/800 (21.8\%)
& 772/800 (96.5\%)
& 86.9 / 86.9
& 598 (0 / 0) \\
\bottomrule
\end{tabular}
\end{table}

\subsection{Impact of the Intervention Strength Scale $\lambda_{\alpha}$}
\label{app:alpha-impact}

\begin{figure}[H]
    \centering
    \includegraphics[width=0.96\linewidth]{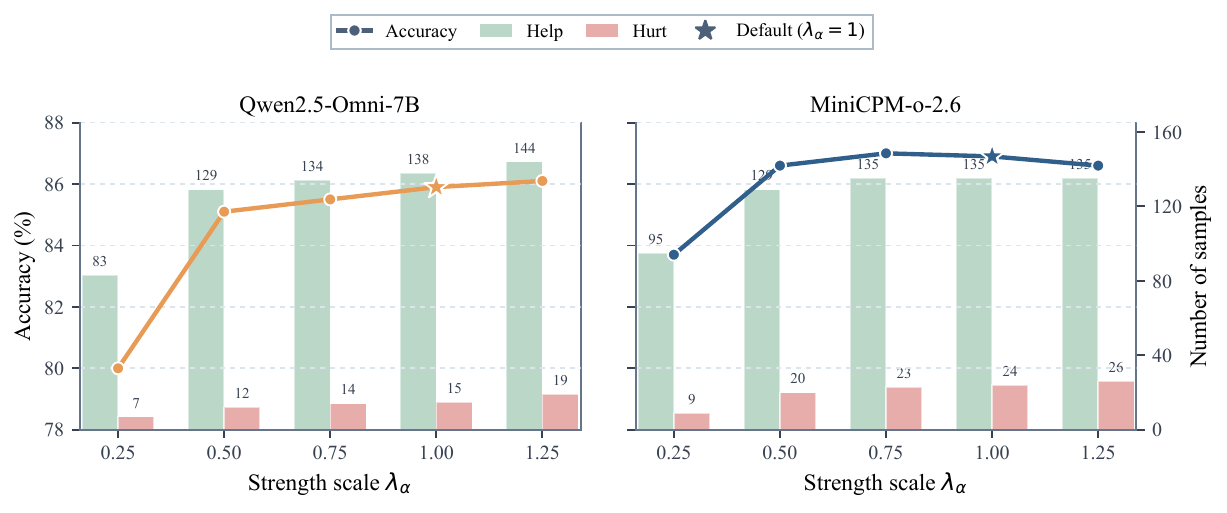}
    \caption{\textbf{Sensitivity to the intervention strength scale $\lambda_{\alpha}$ on CMM.} Lines report accuracy, while bars show the numbers of helpful and harmful prediction changes. Stars denote the default setting $\lambda_{\alpha}=1$.}
    \label{fig:alpha-sensitivity}
\end{figure}

We examine the sensitivity of CaRe to the overall intervention magnitude on CMM. Specifically, after confidence adaptation and depth decay determine the layer-wise coefficient $\alpha_j$, we rescale it as $\widetilde{\alpha}_j=\lambda_{\alpha}\alpha_j$, while keeping all other settings fixed. Figure~\ref{fig:alpha-sensitivity} reports both the resulting accuracy and the numbers of helpful and harmful prediction changes relative to the unmodified backbone.

Performance improves substantially when $\lambda_{\alpha}$ increases from $0.25$ to $0.50$, and remains stable over the moderate range from $0.50$ to $1.25$. Increasing the scale generally produces more helpful corrections, but also introduces more harmful changes, indicating a trade-off between correction capability and over-intervention. The default $\lambda_{\alpha}=1$ achieves near-optimal accuracy on both backbones, falling only 0.2 and 0.1 percentage points below their respective best settings. This result suggests that CaRe is not sensitive to a narrowly tuned intervention scale within the effective range.

\subsection{Impact of the Confidence Threshold $\eta$}
\label{app:eta-impact}

\begin{figure}[h]
    \centering
    \includegraphics[width=0.96\linewidth]{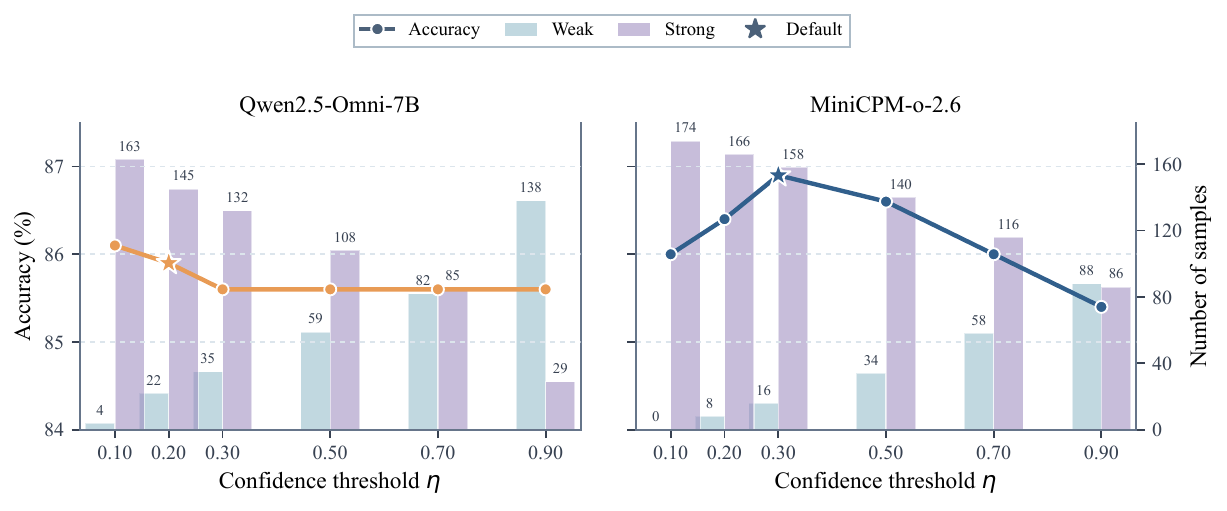}
    \caption{\textbf{Sensitivity to the confidence threshold \(\eta\) on CMM.} Lines report accuracy, while bars show the numbers of samples assigned to the weak and strong intervention regimes. Stars denote the default settings, \(\eta=0.2\) for Qwen2.5-Omni-7B and \(\eta=0.3\) for MiniCPM-o-2.6.}
    \label{fig:confidence_threshold}
\end{figure}

We further examine the sensitivity of CaRe to the confidence threshold \(\eta\), which determines whether a gate-positive sample receives a weak or strong residual intervention. Specifically, samples with counterfactual confidence \(u_r < \eta\) are assigned to the weak regime, while those with \(u_r \geq \eta\) receive the strong intervention. We vary \(\eta\) from \(0.1\) to \(0.9\) while keeping all other hyperparameters fixed. Figure~\ref{fig:confidence_threshold} reports the resulting accuracy together with the numbers of samples assigned to the weak and strong intervention regimes.

As \(\eta\) increases, the intervention allocation shifts monotonically from the strong regime toward the weak regime. For Qwen2.5-Omni-7B, the number of weak interventions increases from 4 at \(\eta=0.1\) to 138 at \(\eta=0.9\), while strong interventions decrease from 163 to 29. A similar trend is observed for MiniCPM-o-2.6, where the weak/strong split changes from \(0/174\) to \(88/86\). This confirms that \(\eta\) directly controls the conservativeness of CaRe: a lower threshold applies strong correction to most conflict-positive samples, whereas a higher threshold reserves strong intervention for only the most confident counterfactual predictions.

The two backbones exhibit different sensitivity patterns. Qwen2.5-Omni-7B achieves its highest accuracy of \(86.1\%\) at \(\eta=0.1\), while the default \(\eta=0.2\) attains \(85.9\%\), only \(0.2\) percentage points lower. Performance becomes nearly unchanged at \(85.6\%\) for larger thresholds. In contrast, MiniCPM-o-2.6 exhibits a clearer intermediate optimum: accuracy increases from \(86.0\%\) at \(\eta=0.1\) to \(86.9\%\) at the default \(\eta=0.3\), and then gradually decreases as the threshold is further increased, reaching approximately \(85.4\%\) at \(\eta=0.9\). These results suggest that overly aggressive strong intervention can be suboptimal when counterfactual confidence is insufficient, whereas an excessively large threshold makes the correction overly conservative and prevents sufficiently confident samples from receiving stronger intervention. Overall, a moderate, backbone-specific confidence threshold provides an effective balance between correction strength and intervention reliability.

\subsection{Joint Sensitivity to $\lambda_{\alpha}$ and $\eta$}
\label{app:joint-sensitivity}

\begin{figure}[h]
    \centering
    \includegraphics[width=0.8\linewidth]{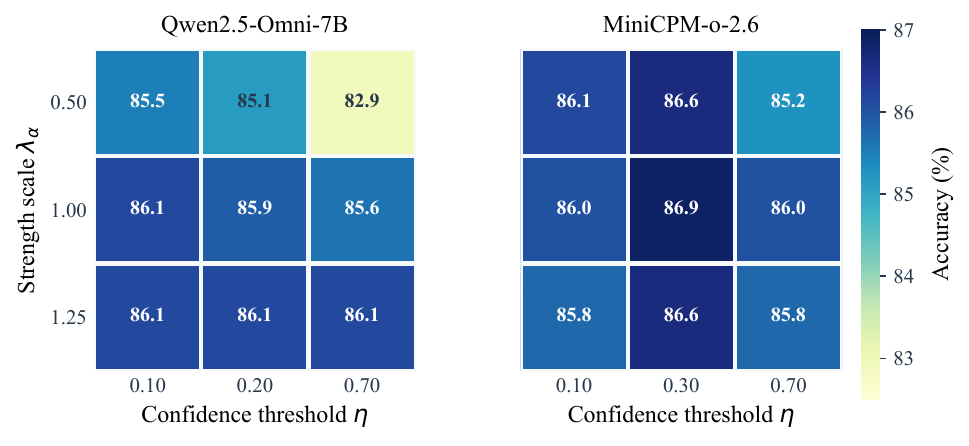}
    \caption{\textbf{Joint sensitivity to the intervention strength scale $\lambda_{\alpha}$ and confidence threshold $\eta$ on CMM.} Each cell reports accuracy (\%); darker colors indicate higher accuracy.}
    \label{fig:lambda-eta-heatmap}
\end{figure}

While $\lambda_{\alpha}$ controls the overall magnitude of residual correction, the confidence threshold $\eta$ determines how frequently the strong intervention regime is activated. We therefore evaluate their joint effect on
CMM using a $3\times3$ grid for each backbone. Figure~\ref{fig:lambda-eta-heatmap} reports the resulting accuracy.

The two hyperparameters exhibit a clear interaction. For Qwen2.5-Omni-7B, a small strength scale becomes sensitive to the threshold: performance drops to 82.875\% when $\lambda_{\alpha}=0.5$ and $\eta=0.7$. Increasing $\lambda_{\alpha}$ mitigates this under-correction, and $\lambda_{\alpha}=1.25$ maintains 86.125\% accuracy across all evaluated thresholds. For MiniCPM-o-2.6, the best result of 86.875\% is obtained at the intermediate configuration $(\lambda_{\alpha},\eta)=(1.0,0.3)$. Increasing the scale to 1.25 does not yield further gains and slightly degrades performance away from this confidence threshold.

Overall, the results show that intervention magnitude and confidence-based allocation should be calibrated jointly rather than independently. The default configurations lie in high-performing regions for both backbones.

\section{Scaling Analysis}
\label{app:scaling}

To further examine whether CaRe generalizes across model capacities, we evaluate Qwen2.5-Omni-3B and Qwen2.5-Omni-7B under the same evaluation protocol. Table~\ref{tab:scaling-qwen3b} reports detailed results on each subset of CMM and AVHBench. CaRe consistently improves overall performance while maintaining gains across different cross-modal hallucination directions.

\begin{table}[t]
\vspace{-0.5em}
\centering
\caption{\textbf{Detailed scaling results on Qwen2.5-Omni-3B (\%).}
We compare the base model with CaRe on CMM and AVHBench.
Results are reported as Accuracy (Acc.) and F1, with \texttt{Yes} treated as the positive class.
\emph{Overall Acc.} is computed over all samples in the corresponding subsets or tasks.
Best results are highlighted in bold.}
\label{tab:scaling-qwen3b}

\vspace{-0.5em}
\small
\setlength{\tabcolsep}{2.1pt}
\renewcommand{\arraystretch}{1.12}

\begin{tabular}{@{}ll cc cc c cc cc c@{}}
\toprule
& & \multicolumn{5}{c}{\textbf{CMM}}
& \multicolumn{5}{c}{\textbf{AVHBench}} \\
\cmidrule(lr){3-7}
\cmidrule(lr){8-12}

\textbf{Backbone} & \textbf{Method}
& \multicolumn{2}{c}{\shortstack[c]{Visual\\Dominance}}
& \multicolumn{2}{c}{\shortstack[c]{Audio\\Dominance}}
& \multirow{2}{*}{\shortstack[c]{Overall\\Acc.}}
& \multicolumn{2}{c}{\shortstack[c]{Video-Driven\\Audio Hall.}}
& \multicolumn{2}{c}{\shortstack[c]{Audio-Driven\\Video Hall.}}
& \multirow{2}{*}{\shortstack[c]{Overall\\Acc.}}
\\

\cmidrule(lr){3-4}
\cmidrule(lr){5-6}
\cmidrule(lr){8-9}
\cmidrule(lr){10-11}

& & Acc. & F1
& Acc. & F1
&
& Acc. & F1
& Acc. & F1
&
\\
\midrule

\multirow{2}{*}{\makecell[c]{Qwen2.5-Omni-3B}}
& Base
& 69.3 & 66.5
& 84.5 & 82.5
& 76.9
& 74.2 & 74.5
& 80.8 & 77.5
& 76.4
\\

& +CaRe
& \textbf{79.8} & \textbf{79.1}
& \textbf{85.3} & \textbf{83.5}
& \textbf{82.5}
& \textbf{81.0} & \textbf{81.4}
& \textbf{80.9} & \textbf{77.6}
& \textbf{81.0}
\\

\bottomrule
\end{tabular}

\vspace{-0.5em}
\end{table}

\section{Reproducibility Details}
\label{app:implementation}

\subsection{CaRe Inference Algorithm}
\label{app:algorithm}

Algorithm~\ref{alg:care} summarizes the complete inference procedure of CaRe. Given an audio input $a$, a visual input $v$, and a textual question $q$, CaRe first performs the original audio-visual inference and uses a text-only modality-demand probe to identify a candidate question-relevant modality. A modality-specific counterfactual branch is constructed only when the probe uniquely favors audio or video. CaRe then compares the prediction of the candidate branch with that of the original joint branch and activates residual correction only when the two branches exhibit a hard decision conflict. Otherwise, the original audio-visual output is retained without modification.

For a gate-positive sample, CaRe computes layer-wise counterfactual residuals between the candidate unimodal and joint audio-visual representations. The decision margin of the candidate branch is converted into a bounded confidence score, which determines whether a weak or strong intervention range is used. Within the predefined intervention layers, the intervention strength decreases with depth, and the residual correction is applied only to the hidden state of the final prompt token. After this prompt-level correction, inference continues along the original audio-visual path with both modalities preserved, and the complete response is generated autoregressively without further intervention.

\begin{algorithm}[t]
\caption{CaRe Inference}
\label{alg:care}
\begin{algorithmic}[1]
\Require Frozen model $f_{\theta}$; audio $a$; video $v$; question $q$;
intervention layers $\mathcal{L}_{\mathrm{int}}=\{l_1,\ldots,l_K\}$;
threshold $\eta$; weak/strong ranges
$[\alpha_{\min}^{W},\alpha_{\max}^{W}]$ and
$[\alpha_{\min}^{S},\alpha_{\max}^{S}]$; decay exponent $\gamma$.
\Ensure CaRe output.

\State Run the joint branch $\mathcal{B}_{AV}=(a,v,q)$ to obtain
$\hat{y}_{AV}$ and $m_{AV}$.
\State Run the text-only modality probe to obtain $(p_a,p_v,p_b)$.

\If{$p_a>\max(p_v,p_b)$}
    \State $r\gets A$, $\mathcal{B}_{r}\gets(a,q)$.
\ElsIf{$p_v>\max(p_a,p_b)$}
    \State $r\gets V$, $\mathcal{B}_{r}\gets(v,q)$.
\Else
    \State \Return original audio-visual output.
\EndIf

\State Run $\mathcal{B}_{r}$ to obtain
$\hat{y}_{r}$, $m_r$, and
$\{h^{r}_{l_j}\}_{j=1}^{K}$.

\If{$m_r=0$ \textbf{or} $m_{AV}=0$
      \textbf{or} $\hat{y}_{r}=\hat{y}_{AV}$}
    \State \Return original audio-visual output.
\EndIf

\State $u_r \gets 2\sigma(m_r)-1$.
\State
$[\alpha_{\min},\alpha_{\max}]
\gets
\begin{cases}
[\alpha_{\min}^{W},\alpha_{\max}^{W}], & u_r<\eta,\\
[\alpha_{\min}^{S},\alpha_{\max}^{S}], & u_r\geq\eta.
\end{cases}$

\State Re-run $\mathcal{B}_{AV}$ for prompt prefilling with
interventions at $\mathcal{L}_{\mathrm{int}}$.

\For{$j=1,\ldots,K$}
    \State At layer $l_j$, let $h^{AV}_{l_j}$ denote the current post-residual hidden state at the final prompt-token position.
    \State $\Delta h^{r}_{l_j}
    \gets h^{r}_{l_j}-h^{AV}_{l_j}$.
    \State $\displaystyle
    \alpha_j\gets
    \alpha_{\min}
    +(\alpha_{\max}-\alpha_{\min})
    \left(\frac{K-j}{K-1}\right)^{\gamma}$.
    \State $\displaystyle
    \widetilde{h}^{AV}_{l_j}
    \gets h^{AV}_{l_j}
    +\alpha_j\Delta h^{r}_{l_j}$.
    \State Propagate $\widetilde{h}^{AV}_{l_j}$ through the
    subsequent decoder layers.
\EndFor

\State Continue autoregressive generation from the intervened
audio-visual prefilling state without further intervention.
\State \Return CaRe output.
\end{algorithmic}
\end{algorithm}

\subsection{Hyperparameter Configurations}
\label{app:hyperparameters}

\begin{table}[t]
\centering
\small
\setlength{\tabcolsep}{4.5pt}
\caption{Hyperparameter configurations of CaRe on CMM and AVHBench.}
\label{tab:care_hyperparams}
\begin{tabular}{llccccc}
\toprule
Backbone & Dataset
& Intervention Layers
& $\gamma$
& $\eta$
& $\alpha^{W}$
& $\alpha^{S}$ \\
\midrule

\multirow{2}{*}{Qwen2.5-Omni-7B}
& CMM
& 25--27
& 0.5
& 0.2
& $[0.2,0.5]$
& $[0.5,0.8]$ \\
& AVHBench
& 25--27
& 0.5
& 0.2
& $[0.2,0.5]$
& $[0.5,0.8]$ \\

\midrule

\multirow{2}{*}{Qwen2.5-Omni-3B}
& CMM
& 33--35
& 0.5
& 0.2
& $[0.2,0.5]$
& $[0.5,0.8]$ \\
& AVHBench
& 33--35
& 0.5
& 0.2
& $[0.2,0.5]$
& $[0.5,0.8]$ \\

\midrule

\multirow{2}{*}{MiniCPM-o-2.6}
& CMM
& 25--27
& 0.5
& 0.3
& $[0.1,0.3]$
& $[0.5,0.8]$ \\
& AVHBench
& 25--27
& 0.5
& 0.5
& $[0.1,0.3]$
& $[0.3,0.5]$ \\

\bottomrule
\end{tabular}
\end{table}

Table~\ref{tab:care_hyperparams} summarizes the hyperparameter configurations used for CaRe across the evaluated settings. Qwen2.5-Omni-7B and MiniCPM-o-2.6 use layers 25--27, while Qwen2.5-Omni-3B uses layers 33--35, corresponding to the final three language-decoder layers of each backbone. We further examine the sensitivity of the main hyperparameters in Appendix~\ref{app:alpha-impact} and Appendix~\ref{app:eta-impact}.

\subsection{Baseline Implementation Details}
\label{app:baselines}

For ICT~\citep{chen2025ict}, AVCD~\citep{jung2026avcd}, and MAD~\citep{chung2026mad}, we strictly adopt their official open-source implementations, directly interfacing them with the omnimodal backbone models. 

For Visual Contrastive Decoding (VCD)~\citep{vcd}, the original method was designed solely for vision-language models by contrasting original representations against visual perturbations. To adapt VCD to the omnimodal setting in a principled and symmetric manner, we implement VCD by contrasting the complete audio-visual input against all possible unimodal and text-only sub-branches, withholding competing sensory inputs rather than injecting heuristic noise. Formally, let $v$ and $a$ represent the visual and audio inputs, and let $\emptyset$ denote a withheld modality. At each decoding step $t$, the calibrated logit $\mathbf{z}_t^{\text{ext}}$ is computed as:
\begin{equation}
    \mathbf{z}_t^{\text{ext}} = \mathbf{z}_t(v, a, q) + \alpha \sum_{(v', a') \in \mathcal{S}} \left( \mathbf{z}_t(v, a, q) - \mathbf{z}_t(v', a', q) \right),
    \label{eq:vcd_ext}
\end{equation}
where $\mathcal{S} = \{(v, \emptyset), (\emptyset, a), (\emptyset, \emptyset)\}$ collects the counterfactual input configurations obtained by withholding the audio input, visual input, or both, respectively.

\section{Prompt Formulations for the Text-Only Modality Probe}
\label{app:probe-prompts}

We provide the exact prompt formulations used in the modality-routing robustness analysis. For all variants, both audio and visual inputs are withheld, and the routing prompt is appended to the original textual question. The placeholder \texttt{\{options\}} is instantiated with an ordered list of \texttt{audio}, \texttt{video}, and \texttt{both}. The main experiments use prompt \textbf{D} with the ordering \texttt{audio, video, both}, while the robustness analysis in Section~\ref{sec:ablation} additionally considers alternative prompt formulations and option orderings.

Table~\ref{tab:probe-variants} lists the default prompt \textbf{D} and three alternative formulations, \textbf{N1--N3}, which share the same modality-routing objective.

\begin{table}[t]
\centering
\caption{\textbf{Prompt formulations for the text-only modality probe.} \textbf{D} is used throughout the main experiments, while \textbf{N1--N3} are alternative formulations considered in the robustness analysis.}
\label{tab:probe-variants}
\begin{tabularx}{\linewidth}{@{}c X@{}}
\toprule
\textbf{ID} & \textbf{Prompt template} \\
\midrule

\textbf{D} &
To answer this question, which modality is needed (\{options\}):
\\[2pt]

\textbf{N1} &
Determine the minimum evidence needed to answer the question. Select one modality: \{options\}:
\\[2pt]

\textbf{N2} &
What kind of input is necessary for a correct answer? Choose from: \{options\}:
\\[2pt]

\textbf{N3} &
Classify the question by the information source it requires. Answer with one of: \{options\}:
\\

\bottomrule
\end{tabularx}
\end{table}

\section{Additional Dataset Details}
\label{app:dataset_details}

\subsection{CMM}
\label{app:cmm}

CMM~\citep{cmm} is designed to evaluate hallucinations in multimodal language models arising from over-reliance on unimodal priors and spurious inter-modality correlations. The benchmark contains six subcategories, among which we focus on \textit{Visual Dominance} and \textit{Audio Dominance} to study bidirectional audio-visual interference. In Visual Dominance, salient visual content may cause the model to hallucinate an associated but absent sound event, whereas in Audio Dominance, auditory cues may induce hallucinated visual objects or events. Each subcategory contains 200 multimodal instances, with each instance paired with one positive question about an existent object/event and one negative question about a non-existent one, yielding 400 binary Yes--No questions per subcategory. We therefore evaluate on 800 questions in total across the two dominance subsets. 

\subsection{AVHBench}
\label{app:avhbench}

AVHBench~\citep{sung2025avhbench} is an audio-visual hallucination benchmark containing 2,136 videos and four tasks covering cross-modal judgment and description. We use its two cross-modal hallucination tasks: \textit{Audio-driven Video Hallucination} and \textit{Video-driven Audio Hallucination}. The former evaluates whether auditory evidence causes a model to falsely perceive an object or event as visually present, while the latter examines whether visible content induces a hallucinated sound event. Both tasks are formulated as balanced binary Yes--No questions. Audio-driven Video Hallucination contains 1,136 questions (568 Yes and 568 No), while Video-driven Audio Hallucination contains 2,290 questions (1,145 Yes and 1,145 No). These complementary directions allow us to evaluate whether CaRe mitigates cross-modal interference when either audio or visual information acts as the misleading modality.

\subsection{Daily-Omni}
\label{app:daily_omni}

Daily-Omni~\citep{zhou2025daily} is a general audio-visual question-answering benchmark designed to evaluate joint perception and reasoning in daily-life scenarios. It contains 684 videos and 1,197 multiple-choice QA pairs spanning six task categories: \textit{Event Sequence}, \textit{Audio-Visual Event Alignment}, \textit{Inference}, \textit{Reasoning}, \textit{Context Understanding}, and \textit{Comparative Analysis}. The videos are segmented into 30-second and 60-second clips, corresponding to 647 and 550 questions, respectively, enabling evaluation across different temporal contexts. Unlike CMM and AVHBench, which explicitly target hallucination under cross-modal interference, Daily-Omni requires broader integration of audio, visual, and textual information. We use the full benchmark to evaluate whether CaRe preserves general audio-visual understanding beyond hallucination-focused settings, and report both category-wise and duration-wise accuracy.

\section{Computational Analysis}
\label{app:computational}

We analyze the inference-time computational cost of CaRe and compare it with AVCD and MAD.
All latency measurements are obtained on the same 200-example evaluation subset for each backbone
under serial execution. We first report the measured end-to-end latency, then decompose the runtime
of CaRe into its major stages. Finally, we estimate the potential latency reduction under ideal
branch-level parallel execution.

\paragraph{End-to-end latency.}
Table~\ref{tab:runtime-serial} reports the end-to-end wall-clock latency under serial execution.
AVCD achieves the lowest mean latency on both backbones, while CaRe remains substantially faster
than MAD. On Qwen2.5-Omni-7B, CaRe requires 6.70\,s per sample on average, compared with
16.61\,s for MAD, corresponding to a 59.7\% reduction. On MiniCPM-o-2.6, the mean latency
decreases from 7.82\,s with MAD to 4.57\,s with CaRe, a reduction of 41.5\%.
The additional cost of CaRe relative to AVCD arises primarily from the modality-specific
counterfactual inference required for conflict diagnosis.

\begin{table}[t]
\centering
\caption{
\textbf{End-to-end inference latency under serial execution.}
Latency is measured over 200 samples for each backbone.
We report the mean, standard deviation, median, and 95th-percentile (P95) latency in seconds.
}
\label{tab:runtime-serial}
\small
\setlength{\tabcolsep}{7pt}
\begin{tabular}{llrrrr}
\toprule
\textbf{Backbone} & \textbf{Method}
& \textbf{Mean} & \textbf{Std.} & \textbf{Median} & \textbf{P95} \\
\midrule
\multirow[c]{3}{*}{Qwen2.5-Omni-7B}
& CaRe & 6.698 & 4.965 & 5.438 & 15.934 \\
& MAD  & 16.612 & 25.150 & 8.797 & 50.130 \\
& AVCD & 4.838 & 2.852 & 4.262 & 11.022 \\
\midrule
\multirow[c]{3}{*}{MiniCPM-o-2.6}
& CaRe & 4.571 & 4.053 & 3.416 & 12.075 \\
& MAD  & 7.816 & 4.157 & 6.245 & 16.573 \\
& AVCD & 2.966 & 1.973 & 2.142 & 7.636 \\
\bottomrule
\end{tabular}
\end{table}

\paragraph{Runtime decomposition of CaRe.}
To identify the source of CaRe's computational overhead, Table~\ref{tab:care-runtime-breakdown}
decomposes its average per-sample latency into the text-only modality probe, the original
audio-visual inference, the selected modality-specific counterfactual inference, the gating and
control logic, and the residual-correction stage.
The original audio-visual and counterfactual branches account for most of the runtime, whereas
the text-only probe and conflict-gating logic introduce negligible overhead.
The residual-correction stage also contributes only a small fraction of the total latency.
In particular, its average cost is 0.305\,s on Qwen2.5-Omni-7B and 0.040\,s on MiniCPM-o-2.6.

\begin{table}[t]
\centering
\caption{
\textbf{Average per-sample latency breakdown of CaRe.}
All values are reported in seconds and averaged over the same 200-example evaluation subset.
\textit{Counterfactual} denotes inference with the selected modality-specific branch, and
\textit{Residual Corr.} denotes the live residual-intervention stage.
}
\label{tab:care-runtime-breakdown}
\small
\setlength{\tabcolsep}{5pt}
\begin{tabular}{lrrrrr}
\toprule
\textbf{Backbone}
& \textbf{Probe}
& \textbf{Joint AV}
& \textbf{Counterfactual}
& \textbf{Gate/Control}
& \textbf{Residual Corr.} \\
\midrule
Qwen2.5-Omni-7B
& 0.021 & 4.369 & 2.004 & $<0.001$ & 0.305 \\
MiniCPM-o-2.6
& 0.021 & 3.031 & 1.479 & $<0.001$ & 0.040 \\
\bottomrule
\end{tabular}
\end{table}

\begin{table}[H]
\centering
\caption{
\textbf{Theoretical latency under ideal branch-level parallel execution.}
Serial latency is measured directly, while parallel latency is estimated from the measured
stage-wise runtimes assuming independent branches can execute concurrently without scheduling,
communication, or device-synchronization overhead. Reduction is relative to serial execution.
}
\label{tab:runtime-parallel}
\small
\setlength{\tabcolsep}{7pt}
\begin{tabular}{llrrr}
\toprule
\textbf{Backbone} & \textbf{Method}
& \textbf{Serial}
& \textbf{Parallel}
& \textbf{Reduction} \\
\midrule
\multirow[c]{3}{*}{Qwen2.5-Omni-7B}
& CaRe & 6.698 & 4.782 & 28.61\% \\
& MAD  & 16.612 & 12.038 & 27.53\% \\
& AVCD & 4.838 & 4.268 & 11.80\% \\
\midrule
\multirow[c]{3}{*}{MiniCPM-o-2.6}
& CaRe & 4.571 & 3.239 & 29.15\% \\
& MAD  & 7.816 & 5.374 & 31.25\% \\
& AVCD & 2.966 & 2.910 & 1.89\% \\
\bottomrule
\end{tabular}
\end{table}

\paragraph{Potential for parallel execution.}
A substantial portion of the serial cost comes from inference branches that need not be executed
strictly sequentially. In CaRe, the modality probe is lightweight and determines which
modality-specific counterfactual branch is required. Once the candidate modality is identified,
the selected counterfactual computation can overlap with the original audio-visual inference,
providing natural branch-level parallelism before the conflict decision and residual correction.
We therefore estimate the theoretical latency obtainable under ideal parallel execution using the
measured stage-wise runtimes.

As shown in Table~\ref{tab:runtime-parallel}, the estimated latency of CaRe decreases from
6.70\,s to 4.78\,s on Qwen2.5-Omni-7B and from 4.57\,s to 3.24\,s on MiniCPM-o-2.6,
corresponding to reductions of 28.6\% and 29.2\%, respectively. This suggests that a substantial
fraction of CaRe's serial overhead originates from independently executable diagnostic computation,
rather than from the residual intervention itself. MAD exhibits a similar potential benefit from
parallel branch execution, whereas the estimated reduction for AVCD is smaller.

\end{document}